%% file: main.tex
\documentclass[runningheads]{llncs}
\usepackage{graphicx}
\usepackage{subcaption}

\usepackage[utf8]{inputenc} 
\usepackage[T1]{fontenc}    
\usepackage{hyperref}       
\usepackage{url}            
\usepackage{booktabs}       
\usepackage{amsfonts}       
\usepackage{nicefrac}       
\usepackage{microtype}      
\usepackage{xcolor}         

\usepackage{graphicx}
\usepackage{subcaption}
\usepackage{tikz}
\usepackage{comment}
\usepackage{amsmath,amssymb} 
\usepackage{color}
\usepackage{siunitx,caption}
\usepackage{subcaption}
\usepackage{pifont}
\usepackage{multirow}
\usepackage{makecell}
\usepackage{bold-extra}
\usepackage{listings}
\usepackage{algorithm}

\usepackage[accsupp]{axessibility}  

\newcommand{\ModelName}{\textsc{Tall}Former}
\newcommand{\cmark}{\ding{51}}%
\newcommand{\xmark}{\ding{55}}%

\begin{document}


\pagestyle{headings}
\mainmatter
\def\ECCVSubNumber{1084}  

\title{\ModelName: Temporal Action Localization with a Long-memory Transformer} 

\titlerunning{\ModelName: TAL with a Long-memory Transformer}
%
\authorrunning{F. Cheng, G. Bertasius}

\author{Feng Cheng\inst{1} \qquad Gedas Bertasius\inst{1} \\
    \texttt{\{fengchan,gedas\}@cs.unc.edu}
}
\institute{Department of Computer Science, University of North Carolina at Chapel Hill}



\title{\ModelName: Temporal Action Localization with a Long-memory Transformer} 

\maketitle

\begin{abstract}
Most modern approaches in temporal action localization divide this problem into two parts: (i) short-term feature extraction and (ii) long-range temporal boundary localization. Due to the high GPU memory cost caused by processing long untrimmed videos, many methods sacrifice the representational power of the short-term feature extractor by either freezing the backbone or using a small spatial video resolution. This issue becomes even worse with the recent video transformer models, many of which have quadratic memory complexity. To address these issues, we propose \textsc{Tall}Former, a memory-efficient and end-to-end trainable \textbf{T}emporal \textbf{A}ction \textbf{L}ocalization transformer with \textbf{L}ong-term memory. Our long-term memory mechanism eliminates the need for processing hundreds of redundant video frames during each training iteration, thus, significantly reducing the GPU memory consumption and training time. These efficiency savings allow us (i) to use a powerful video transformer feature extractor without freezing the backbone or reducing the spatial video resolution, while (ii) also maintaining long-range temporal boundary localization capability.  With only RGB frames as input and no external action recognition classifier, \ModelName~outperforms previous state-of-the-arts by a large margin, achieving an average mAP of 59.1\% on THUMOS14 and 35.6\% on ActivityNet-1.3. The code is public available\footnote{\url{https://github.com/klauscc/TALLFormer}.}

\end{abstract}

\input{introduction}

\input{related_works}
\input{our_approach}

\input{experiments}
\input{conclusions}

%
\bibliographystyle{splncs04}
\bibliography{egbib}

\appendix
\input{appendix}

\end{document}

%% file: introduction.tex
\section{Introduction}

With the rapid growth of video media, video understanding has become an important area of computer vision. As a fundamental task in video understanding, Temporal Action Localization (TAL) aims to localize temporal boundaries and classify the actions for each action instance in a long untrimmed video.

\begin{figure}[t]
    \centering
    \includegraphics[width=0.75\textwidth]{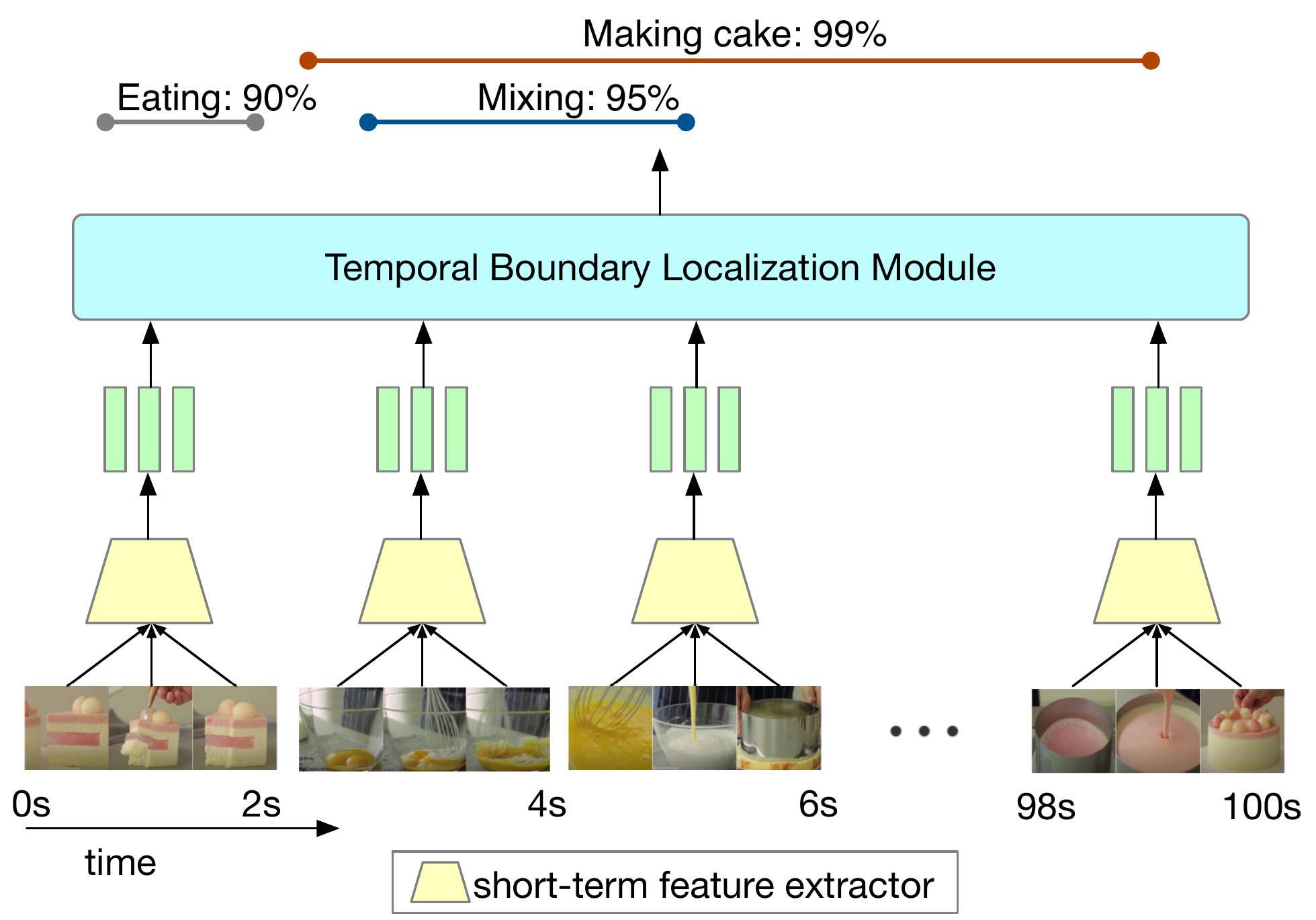}
    \captionsetup{font=small}
    \caption{A general framework for temporal action localization (TAL). The short-term feature extractor extracts the features for each short-term clip. Then, the long-term temporal boundary localization module uses the features of all short-term clips in the video to predict the action boundaries and categories. Due to excessive training time and GPU memory cost, many prior TAL methods restrict the representational power of the short-term feature extractor by either freezing the backbone or operating on small spatial video resolution. While effective at reducing the computational burden, these techniques also significantly degrade TAL performance.}
    \label{fig:long_short_modeling}
\end{figure}

Because many actions span long temporal extent (e.g., 50-100s), most prior approaches in TAL~\cite{lin2018bsn,lin2019bmn,xu2020g,chao2018rethinking,bai2020boundary,zhao2020bottom,long2019gaussian,lin2021learning,wang2021rgb}, divide this problem into two parts: (i) short-term feature extraction and (ii) long-range temporal boundary localization. As shown in Fig.~\ref{fig:long_short_modeling}, the first part involves sampling many consecutive short clips (e.g., each spanning 1-2 seconds)  from a long untrimmed video and extracting short-term features from them. In the second part, the model uses the extracted features of all short-term clips (i.e., spanning the entire duration of an untrimmed video) for predicting action boundaries and categories. Thus, based on these observations, it is natural to conclude that an ideal TAL model should consist of (i) a powerful short-term feature extractor and (ii) a precise temporal boundary localization module.

However, due to the high GPU memory cost needed to process long untrimmed videos, the majority of existing methods sacrifice the representational power of short-term feature extractor, by either freezing the backbone \cite{lin2018bsn,lin2019bmn,xu2020g} or using a very small spatial video resolution (e.g., $96\times96$) \cite{lin2021learning,wang2021rgb}. While both of these techniques are highly effective at reducing GPU memory consumption, they also degrade the quality of extracted short-term features, which leads to a significantly lower TAL accuracy.
For example, as shown in Table~\ref{tab:feature_extraction}, in order to save memory, reducing spatial resolution from $168\times 168$ to $112\times 112$ leads to 2.3\% mAP drop; using a smaller backbone also reduces the mAP by 2.1\%; freezing the backbone leads to a severe drop in mAP ($\sim$8-9\%).

\begin{table}[htb!]
\small
 \captionsetup{font=small}
 \caption{We study several important factors for short-term feature extraction on THUMOS14: (i) spatial video resolution, (ii) transformer backbone complexity, and (iii) the number of frozen backbone stages. For these experiments, we use Swin~\cite{liu2021video}, which consists of 4 stages where $i$ frozen stages means that the first $i$ stages in the backbone are frozen. We use DaoTAD~\cite{wang2021rgb} codebase to conduct these experiments. Additionally, we note that in all of these experiments, we incorporate Checkpointing~\cite{chen2016training} to reduce GPU memory usage. GPU Memory (Mem) is measured in Gigabytes. Based on these results, we note that \textbf{(a)} increasing the spatial resolution leads to large mAP improvement($\sim$2.3\%) but also quadratic memory consumption, \textbf{(b)} using larger backbones also improves the mAP, and lastly, \textbf{(c)} freezing the backbone leads to a severe drop in performance ($\sim$8-9\%).} 
    \centering
    \begin{subtable}[t]{.32\linewidth}
         \label{tab:feat_extract-sr}
        \centering
        \captionsetup{font=footnotesize, width=0.8\textwidth}
        \caption{Spatial video resolution (SVR) analysis. The backbone is Swin-T with the first 2 stages frozen.}
        \begin{tabular}{ccc}
        \toprule
        SVR & mAP(\%) & Mem \\
        \midrule
        $112\times112$ & 52.8 & \textbf{15}  \\
        $168\times168$ & \textbf{55.1} & 34 \\
        $224\times224$ & - & OOM \\
        \bottomrule
        \end{tabular}
    \end{subtable}
    \hspace{4pt}
    \begin{subtable}[t]{.32\linewidth}
         \label{tab:feat_extract-backbone}
        \centering
        \captionsetup{font=footnotesize, width=0.9\textwidth}
        \caption{Studying the backbone complexity. Spatial resolution is $112\times 112$. The first 2 stages of backbone are frozen.}
        \begin{tabular}{ccc}
        \toprule
        Backbone & mAP(\%) & Mem \\
        \midrule
        Swin-T & 52.8 & \textbf{15} \\
        Swin-S & 53.3 & 17 \\
        Swin-B & \textbf{54.7}  & 20 \\
        \bottomrule
        \end{tabular}
    \end{subtable}
    \begin{subtable}[t]{.32\linewidth}
         \label{tab:feat_extract-frozen_stages}
        \centering
        \captionsetup{font=footnotesize, width=0.8\textwidth}
        \caption{The number of frozen backbone stages (FBS). Spatial resolution is $112 \times 112$. The backbone is Swin-T.}
        \begin{tabular}{ccc}
        \toprule
        FBS & mAP(\%) & Mem \\
        \midrule
        4 & 44.3 & \textbf{3} \\
        2 & 52.8 & 14 \\
        0 & \textbf{53.8} & 26 \\
        \bottomrule
        \end{tabular}
    \end{subtable}
    \label{tab:feature_extraction}
\end{table}

In parallel, we note that the recent introduction of powerful video transformer models~\cite{bertasius2021space,liu2021video} have achieved impressive results on various video understanding problems such as action recognition. However, these models have made the above-described GPU memory issues even worse. Due to the quadratic complexity of self-attention, video transformers require even more GPU memory than traditional CNNs. As a result, it is challenging to adapt these models to the TAL task, which generally requires a lot of GPU memory even when using CNN-based models. The commonly used GPU memory saving techniques such as Checkpointing~\cite{chen2016training}, Mixed Precision~\cite{micikevicius2017mixed}, can alleviate these computational issues. However, as shown in Table~\ref{tab:feature_extraction}, even when using these techniques, the GPU memory cost of applying video transformers (e.g., VideoSwin~\cite{liu2021video}) to TAL is very large.

Thus, with these computational issues in mind, we propose \ModelName, a memory-efficient and end-to-end trainable \textbf{T}emporal \textbf{A}ction \textbf{L}ocalization Transformer with a \textbf{L}ong-memory mechanism. Our key observation is that most videos are highly redundant, i.e., their content changes little in most neighboring frames. This raises the question of whether every single frame from a long untrimmed video needs to be processed during each training iteration. Motivated by this observation, we design \ModelName~ to process only a fraction of randomly selected frames at each training iteration, which significantly reduces the training time and GPU memory requirements. For the remaining (i.e., not selected) video frames, the video features are sampled from  long-term memory, which stores the features of all previously processed frames for that particular video. Note that the features from long-term memory do not have to be re-computed online, and they also do not require backpropagating the gradients, which makes long video processing much more efficient.  

As the short-term feature extractor evolves throughout training, the video features in long-term memory are also evolving, i.e., the newly computed features for a given video are used to replace the old features in long-term memory. Compared to previous TAL approaches, \textsc{Tall}Former has several main advantages. First, our model can be trained end-to-end on long, high spatial resolution videos beyond the constraints of finite GPU memory. Second, our framework is flexible as we can incorporate any state-of-the-art short-term video transformer model into \textsc{Tall}Former, thus, benefiting from future improvements in the video transformer design. Lastly, unlike many previous TAL methods~\cite{lin2018bsn,lin2019bmn,xu2020g,bai2020boundary,zhao2020bottom,long2019gaussian} that rely on external action recognition classifiers, \textsc{Tall}Former is a unified framework that predicts action boundaries and categories with a single model. Despite being simpler, and only operating on RGB inputs, \textsc{Tall}Former achieves  an average mAP of 59.1\% on THUMOS14 and 35.6\% on ActivityNet-1.3, thus, outperforming the current state-of-the-arts by 7.1\% and 1.2\% respectively.

%% file: related_works.tex
\section{Related Work}

\paragraph{Action Recognition.}
Action recognition is a fundamental short-term modeling task in video understanding. With the success of deep learning, a vast array of methods \cite{tran2015learning,wang2016temporal,tran2018closer,wang2018appearance,carreira2017quo,feichtenhofer2020x3d,feichtenhofer2019slowfast,jiang2019stm,lin2019tsm,kwon2020motionsqueeze,wang2018temporal,you2022megan} utilize 2D and 3D CNNs to achieve impressive performance on standard action recognition benchmarks~\cite{carreira2017quo}. Recently, Vision Transformer-based methods \cite{bertasius2021space,liu2021video,fan2021multiscale,you2022class} have been shown to outperform previous CNN-based methods by a large margin.  Due to the large scale pretraining on action recognition datasets, the pretrained models from this domain are widely used in temporal action localization as a short-term feature extractor. One limitation of modern video transformer models is that due to the quadratic memory complexity of self-attention \cite{vaswani2017attention}, these models are slow to train and they require a lot of GPU memory. As a result, it is difficult to apply them to long-term modeling tasks such as temporal action localization.

\paragraph{Temporal Action Localization (TAL).}

Due to finite GPU memory constraints, most existing methods \cite{lin2018bsn,lin2019bmn,xu2020g,chao2018rethinking,bai2020boundary,zhao2020bottom,long2019gaussian,tan2021relaxed,zhao2021video,bagchi2021hear} use pre-extracted action recognition features as inputs to the TAL model. However, since those features are extracted using models~\cite{he2016deep,carreira2017quo} that are pretrained on different datasets, using these features for TAL  often leads to suboptimal performance. To address these issues, recent methods AFSD \cite{lin2021learning} and DaoTAD \cite{wang2021rgb} proposed end-to-end trainable frameworks. However, to fit into finite GPU memory, these models operate on very low spatial video resolutions (e.g., $96 \times 96$ and $112 \times 112$ respectively), which leads to a significant drop in TAL accuracy. To the best of our knowledge, none of the existing methods are capable of end-to-end training with both high spatial resolution and long temporal extent. We aim to address this issue by proposing a simple, end-to-end trainable, transformer-based TAL method that can operate on long high-resolution video inputs.

Besides end-to-end training ability, we also note that most TAL methods can be categorized into two groups: (i) single-stage detectors, and (ii) two-stage detectors that require external action recognition classifiers. One-stage detectors \cite{lin2017single,liu2020progressive,zhang2018s3d,wang2020multi} perform action localization and classification at the same time. In comparison, the two-stage methods \cite{lin2018bsn,lin2019bmn,xu2020g,lin2021learning,gao2018ctap,zeng2019graph,liu2019multi,gao2020accurate,bai2020boundary,su2020bsn++,qing2021temporal,shou2017cdc,zhao2017temporal} only predict action boundaries and then use the predictions of an external action recognition classifier to assign an action class to a given video segment.
Despite the elegance and simplicity of one-stage methods, the two-stage methods typically have a much higher detection accuracy. 
In this work, we will show that even without relying on the external action recognition classifier, our \ModelName~still achieves state-of-the-art results on several major TAL benchmarks. 

\paragraph{Memory-saving Techniques.}
Applying transformer-based methods to TAL poses many GPU memory challenges due to the quadratic memory complexity of self-attention. There are several general memory-saving techniques, including Gradient Checkpointing~\cite{chen2016training} and Mixed Precision~\cite {micikevicius2017mixed}, which reduce the GPU memory usage by about 50\%. We note that our proposed approach is complementary to these techniques. In fact, we use Gradient Checkpointing~\cite{chen2016training} in many of our experiments, thus, demonstrating that our proposed method works well in conjunction with these prior memory-saving techniques. 

Furthermore, we note that several methods from Natural Language Processing (NLP) such as LinFormer~\cite{wang2020linformer} and Performer~\cite{choromanski2020rethinking} propose to reduce the memory complexity of standard self-attention by approximating the attention using low-rank matrix decomposition. While being effective in NLP, those approximation methods work poorly when applied to video recognition~\cite{patrick2021keeping}. 

%% file: our_approach.tex
\section{\ModelName}

\definecolor{codegreen}{rgb}{0,0.6,0}
\definecolor{codegray}{rgb}{0.5,0.5,0.5}
\definecolor{codepurple}{rgb}{0.58,0,0.82}
\definecolor{backcolour}{rgb}{0.95,0.95,0.92}

\begin{figure}[t]
    \centering
    \includegraphics[width=0.9\textwidth]{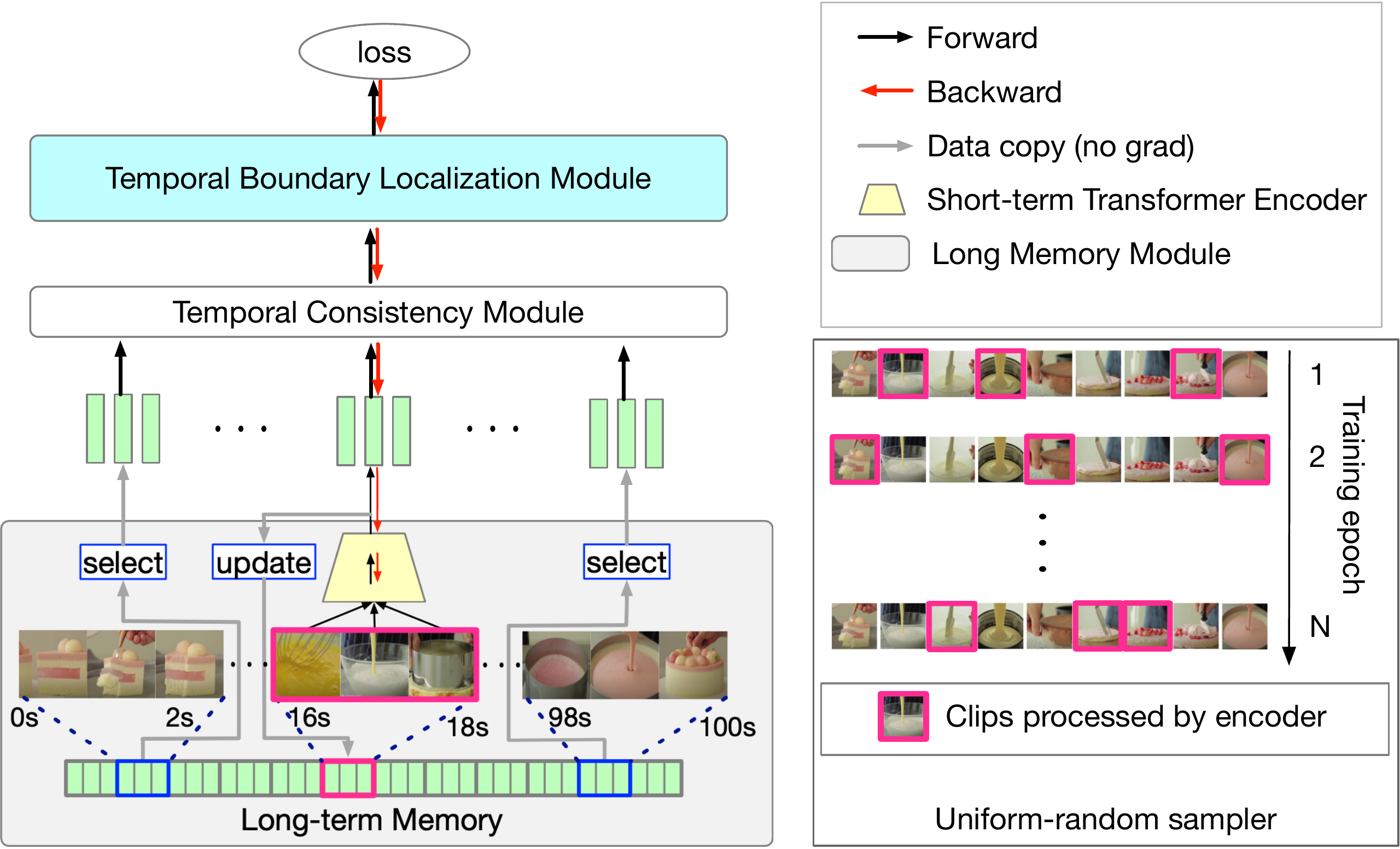}
    \captionsetup{font=small}
    \caption{An illustration of our proposed \ModelName~model. Our method consists of four high-level components: (i) a short-term Transformer encoder, (ii) a long memory module, (iii) a temporal consistency module and (iv) a temporal boundary localization module. The short-term Transformer encoder only extracts features from a few randomly sampled video clips. The rest of the features are sampled from long-term memory. All features are then fed into a temporal consistency module to ensure smooth feature fusion. Lastly, the temporal boundary localization module outputs temporal boundaries and action categories for each action instance. Afterward, the features extracted by the short-term encoder are used to update the corresponding features in long-term memory.
    }
    \label{fig:tallformer}
\end{figure}

Given an untrimmed video $V = \{x_t\}_{t=1}^{T} \in \mathbb{R}^{C\times T \times H \times W}$ with $T$ RGB frames, our \ModelName~model aims to predict a set of action instances $\Phi_{V} = \{\phi_m\}_{m=1}^{M}$ where $M$ is the number of action instances in $V$. Each action instance $\phi_m = (s_m, e_m, c_m, p_m)$ is a four-element tuple that represents the start timestamp of action, end timestamp of action, action class and probability of this instance respectively.

As shown in Fig.~\ref{fig:tallformer}, \ModelName~consists of four components: (i) a short-term Transformer encoder, (ii) a long memory module, (iii) a temporal consistency module and (iv) a temporal boundary localization module. First, we randomly sample a subset of short video clips, and process them using the short-term Transformer encoder. The remaining features are directly sampled from long-term memory, which stores previously computed features of all frames for that particular video input. Afterward, all of these features (i.e., from the short-term Transformer encoder and  long-term memory) are fed into a temporal consistency module that effectively fuses them in order to map them to a similar feature space, i.e., to alleviate potential issues caused by differing feature distributions from the feature extractor and long-term memory. Lastly, the temporal boundary localization module processes these features and produces temporal boundaries and action categories for each detected action instance.
We now describe each of these components in more detail.

\subsection{Short-term Transformer Encoder}
\label{sec:st_extractor}

Our Shor-term Transformer Encoder considers many consecutive short clips (i.e., spanning 1-2 seconds) from a long untrimmed video. In order to avoid computing dense features for every single clip, we randomly sample a fixed number of such clips and feed them into our encoder. 

Formally, for each input video $V$ we divide it into $N_c$ non-overlapping clips $c = \{c_m\}_{m=1}^{N_c}$ where $c_m \in \mathbb{R}^{L_c \times H \times W \times 3} $.
The input video is shifted at most $L_c$ frames to ensure the clip-division changes at each epoch.
A uniform sampler first samples the indices $I \in \mathbb{R}^{N_s}$ of clips that will be processed by the encoder. The indices of the remaining (i.e., not sampled) clips are denoted as $I' \in \mathbb{R}^{N_c-N_s}$. The encoder then processes each sampled clip $c_{i}$ to extract low-dimensional features $f_{c_i} \in \mathbb{R}^{L_f\times C_f}$ to produce features $f_I^{(s)} = \{f_{c_{I_1}}, f_{c_{I_2}}, ..., f_{c_{I_{N_s}}}\} \in \mathbb{R}^{N_s \times L_f \times C_f}$.

Note that during each training iteration, the Transformer encoder only processes a small fraction of clips from the whole input video. The remaining clips are sampled from the long memory module (described in Sec.~\ref{sec:lmm}). This enables \ModelName~to be trained end-to-end on long high spatial resolution videos without (i) reducing the spatial video resolution, (ii) freezing the backbone, or (iii) resorting to a weak short-term feature extraction backbone. We use the recent VideoSwin~\cite{liu2021video} as our short-term Transformer encoder, which achieved impressive results on several popular action recognition benchmarks~\cite{carreira2017quo,goyal2017something}.

\subsection{Long Memory Module}
\label{sec:lmm}

Our proposed Long Memory Module (LMM) enables \ModelName~to be trained on long and high-resolution videos. Inspired by~\cite{cheng2022stochastic,zhang2021temporal}, we propose LMM to cache the features computed by our short-term Transformer encoder for all short-term video clips.
For the remaining clips (denoted by the indices $I'$) that are not processed by the short-term Transformer encoder, LMM samples the features $f_{I'}^{(l)} \in \mathbb{R}^{(N_c-N_s)\times L_f \times C_f}$ from long-term memory.
Following this step, we then update long-term memory with the features $f_{I}^{(s)}$ extracted by the short-term Transformer encoder. Note that before training, we initialize the LMM with the features extracted by our short-term Transformer encoder.  

Such a scheme works well in the TAL setting because the short-term Transformer encoder is already pretrained on a large-scale external action recognition dataset (e.g., Kinetics) and thus, it evolves more slowly than the other modules in the network (i.e., it uses a smaller learning rate than the other parts of the network). Thus, ``approximating" short-term features with the features from LMM provides large efficiency gains (both in terms of training time and GPU memory), while still achieving excellent TAL accuracy, which we demonstrate in our experimental section.
Compared to prior methods~\cite{wu2019long,he2020momentum,xu2021long,zhang2021temporal} that use memory bank as auxiliary information, our LMM serves as an approximation to the short-term encoder. Both the features from LMM and short-term encoder are directly used to produce the final predictions of our method.

Overall, compared to standard end-to-end training, \ModelName~only needs to process a fraction of input clips, which saves the memory and computational cost by a rate of $r = \frac{N_s}{N_c}$. This then allows us to (i) use a powerful transformer-based feature extractor without freezing its backbone or reducing the spatial video resolution and (ii) still maintain the ability to precisely localize long-range temporal boundaries of actions. Note that during inference, we extract all features using a short-term Transformer encoder (i.e., without using LMM).

\subsection{Temporal Consistency Module}
Due to different feature distributions between (i) the online extracted Transformer features $f_{I}^{(s)}$ and (ii) LMM-cached offline features  $f_{I'}^{(l)}$, we need to reduce temporal inconsistency among clip-level features across the whole input video. To be more precise, the features that are processed online (i.e., using our short-term Transformer encoder) are extracted using the latest short-term encoder. In contrast, most clip-level features stored in the LMM are extracted using the same short-term Transformer encoder but from the previous iterations. Thus, the short-term features associated with different clips might have different feature distributions, which can potentially degrade TAL performance. To address this issue, we propose a simple, yet effective Temporal Consistency Module (TCM).

The idea is to make the features from both sources more consistent by allowing them to interact with each other. Due to the effectiveness of standard self-attention to capture global long-range dependencies, we design TCM as an $L$ attention layer subnetwork. Formally, given the video features $g = [f^{(l)}_I; f^{(s)}_{I'}]$, the TCM refines the features using three Transformer layers:
\begin{equation}
    h^{(i)} = \text{TransformerLayer}(h^{(i-1)})
\end{equation}
where $i \in [1,L]$ is the layer index, $h^{(0)} = g$ and $h^{(L)}$ is the refined features of TCM. The TransformerLayer uses relative positional encoding as in Swin~\cite{liu2021swin}, GELU~\cite{hendrycks2016gaussian} activation, and Droppath~\cite{huang2016deep}. 

Conceptually, our self-attention-based TCM subnetwork allows our model to refine potentially inconsistent features by incorporating temporal information from the entire untrimmed input video into feature vectors associated with individual video clips. In our experimental section, we demonstrate the effectiveness of such long-range TCM module.

\subsection{Temporal Boundary Localization Module}
The Temporal Boundary Localization Module (TBLM) utilizes the features of all clips produced by the TCM to predict the action boundaries and categories.
The TBLMs in most existing methods~\cite{lin2018bsn,lin2019bmn,xu2020g,bai2020boundary,zhao2020bottom,long2019gaussian,lin2021learning,tan2021relaxed,zhao2021video} especially for difficult datasets (e.g., ActivityNet~\cite{caba2015activitynet}) are two-stage detectors that require external action classification classifiers, which is costly and cumbersome.
Our analysis into this problem reveals that the reason that many prior methods rely on external classifiers is because of a weak short-term encoder. Specifically, as discussed above, many prior methods have to either freeze the backbone or use small spatial video resolution in order to save GPU memory. This then leads to poor action classification performance, which requires these methods to adapt an external action recognition classifier. In contrast, we note that \ModelName~utilizes a strong short-term encoder while achieving strong performance on both action classification and localization using a one-stage TBLM.

We build upon the existing methods~\cite{lin2021learning,wang2021rgb} by simply adding a shared linear action recognition classifier to each action proposal. 
For datasets where each video only contains one action category such as ActivityNet~\cite{caba2015activitynet},
we add the linear classifier on the temporally averaged features of the TCM with a 50\% dropout. Due to the strong representational power of \ModelName, this simple modification achieves a high action classification accuracy and thus, eliminates the necessity for an external action recognition classifier.
We use the same loss functions as the previous methods~\cite{lin2021learning,wang2021rgb} and Focal Loss~\cite{lin2017focal} for the added linear action recognition layer. See more details in Appendix. \ref{a_tblm}.

%% file: experiments.tex
\section{Experiments}
\subsection{Datasets and Evaluation Metrics}
\hspace{\parindent}\textbf{Datasets.} We conduct our evaluations on the two commonly-used benchmark datasets THUMOS14 \cite{idrees2017thumos}, and ActivityNet-1.3 \cite{caba2015activitynet}. THUMOS14 contains 200 untrimmed validation videos and 213 untrimmed testing videos with temporal annotations from 20 categories. ActivityNet-1.3 contains $15,000$ videos for training and $5,000$ videos for validation. Additionally, we also evaluate on the large-scale HACS-Segment~\cite{zhao2019hacs} dataset, which contains 38K untrimmed videos for training and 6K for validation.
Following previous works \cite{lin2019bmn,lin2021learning,wang2021rgb}, on THUMOS14, we train on the validation set and evaluate on the test set. On ActivityNet-1.3 and HACS-Segment we train on the training set and evaluate on the validation set.

\textbf{Evaluation Metrics.} As is standard,  we use mean Average Precision (mAP) to report our results. The Intersection over Union (IoU) thresholds are set to $[0.3:0.1:0.7]$ for THUMOS14 and $[0.5:0.05:0.95]$ for ActivityNet-1.3 and HACS-Segment.

\subsection{Implementation Details}

The flexibility of our framework allows us to consider any transformer-based model as our short-term feature extractor. Due to its superior accuracy, we adopt Video Swin Transformer \cite{liu2021video} pretrained on Kinetics-400 \cite{carreira2017quo}.
The number of layers $L$ in TCM is set to 3 and Droppath rate is 0.1.
Our temporal boundary localization module (TBLM) is designed using the techniques from DaoTAD \cite{wang2021rgb} and AFSD \cite{lin2021learning} for THUMOS14 and ActivityNet-1.3 respectively. Unlike previous methods that operate on (i) RGB and (ii) optical flow frames inputs, our \ModelName~only uses RGB frames. The frames are resized to $256\times 256$ and cropped to $224\times 224$ unless stated otherwise.

During training, we apply common data augmentations on both datasets, including random crop, random horizontal flipping, random rotate and other photometric distortions such as random brightness and contrast.
For the other training and inference details, we follow DaoTAD for THUMOS14 and AFSD for ActivityNet-1.3 with minor modifications as below. The inference and all the other settings are kept the same. Gradient Checkpointing~\cite{chen2016training} is applied to all our models. Our models are trained on 4$\times$ RTX A6000 GPUs.

\textbf{THUMOS14.} We extract RGB frames with 15fps. Because 99.5\% of action instances in the validation set span less than 32 seconds, we consider 480 frames as inputs. The batch size is set to 4 on each GPU. Since in DaoTAD, clip features are temporally downsampled by 8, but in Swin Transformer the temporal downsampling rate is only 2, we add two Convolutional layers with stride 2 before the TCM to keep the same temporal downsampling rate as in DaoTAD.

\textbf{ActivityNet-1.3.} As is done in prior work~\cite{lin2021learning}, we resize all videos to 768 RGB frames. The batch size is set to 1 on each GPU. For simplicity, we remove the boundary consistency learning module in AFSD. Our model is trained for 10 epochs instead of 16 as used in the original AFSD.

\subsection{Comparison with Short-term and Long-term Baselines}

We next conduct a thorough empirical study investigating the importance of short-term vs. long-term modeling for the TAL task. We focus our comparisons on three of our baselines, which are compared under the same finite GPU memory constraints, i.e., either $12$GB (RTX 3080) or $32$GB (Tesla V100).

    \textbf{LT-Frozen}: For this Long-Term modeling baseline, we use a powerful yet frozen Video Swin-B as the feature extractor. A similar strategy of freezing the feature extractor is commonly used in many prior methods~\cite{lin2018bsn,lin2019bmn,xu2020g} as the GPU memory savings from freezing the backbone enable long-range temporal modeling needed by TAL. All models are trained under a finite GPU memory constraint.
    
     \textbf{ST-E2E}: Unlike LT-Frozen baseline, the Short-Term End-to-End trainable baseline uses a Swin-B feature extractor (\textit{not frozen}) that operates on $224 \times 224$ video frame inputs. While benefiting from end-to-end trainability, due to the GPU memory limitation, this baseline can only span either (i) short temporal extent with dense video frame sampling or (ii) long temporal extent with sparse video frame sampling. We study both of these ST-E2E variants.
     
     \textbf{\ModelName}: Compared to the previous two baselines, we believe that our approach achieves the best trade-off between short-term and long-term modeling. In other words, the short-term feature extractor in our framework can be trained end-to-end on high spatial resolution videos. Furthermore, our long-term memory module enables the model to maintain strong long-term modeling capability for precise temporal boundary localization.

\begin{table}[tbp]
\setlength{\tabcolsep}{3pt}
\small
    \centering
    \captionsetup{font=small}
    \caption{Comparing \ModelName~with several of our own short-term and long-term baselines on THUMOS14. We use a powerful video Swin-B as the Feature Extractor (\textbf{FE}) for all models. All models operate on videos with a Spatial Resolution (\textbf{SR}) of $224 \times 224$. LT-Frozen is a long-term baseline that uses a frozen feature extractor but long Temporal Support (\textbf{TS}). ST-E2E is a short-term end-to-end trainable baseline with an unfrozen backbone but short temporal support. \ModelName~ provides the best trade-off between short-term and long-term modeling among these baselines.
    }
    \begin{tabular}{c c c c c c c c c c c}
    \toprule
       Mem Cap & Model Type & \multicolumn{3}{c}{Short-term} & \multicolumn{2}{c}{Long-term} & \multicolumn{4}{c}{mAP(\%)} \\
        \cmidrule(lr){3-5} \cmidrule(lr){6-7} \cmidrule(lr){8-11}
        (GB/GPU) & & FE & E2E & SR & TS(s) & Fps & 0.3 & 0.5 & 0.7 & Avg. \\
         \midrule
    \multirow{5}{*}{12} & LT-Frozen & Swin-B & \xmark & 224 & 32 & 15 & 70.2&55.4&29.4&52.7 \\
    \cmidrule(lr){2-11}
                        & \multirow{3}{*}{ST-E2E} &  Swin-B & \cmark & 224 & 8 & 8 & 63.1 &46.6&16.0&42.9 \\
                        &  &  Swin-B & \cmark & 224 & 32 & 2 & 55.4 & 32.2 & 8.4 & 31.9 \\
                        &  &  Swin-B & \cmark & 224 & 4 & 15 & 50.5 & 33.9 & 13.4 & 32.8  \\
    \cmidrule(lr){2-11}
                        & \ModelName &  Swin-B & \cmark & 224 & 32 & 15 & \textbf{76.1} & \textbf{63.1} &\textbf{34.2}&\textbf{59.0}\\
         \midrule
        \multirow{4}{*}{32} & LT-Frozen &  Swin-B & \xmark & 224 & 32 & 15 & 70.2&55.4&29.4&52.7\\
    \cmidrule(lr){2-11}
         &\multirow{2}{*}{ST-E2E} &  Swin-B & \cmark & 224 & 32 & 8 & 72.7&59.8&33.3&56.3\\
         &  &  Swin-B & \cmark & 224 & 12 & 15 & 72.8&58.6&30.0&55.1\\
    \cmidrule(lr){2-11}
         & \ModelName &  Swin-B & \cmark & 224 & 32 & 15 & \textbf{76.0} &\textbf{63.2}&\textbf{34.5}&\textbf{59.2}\\
         
         \bottomrule
    \end{tabular}
    \label{tab:tradeoffs}
\end{table}

\textbf{Analysis.} From Table~\ref{tab:tradeoffs}, we observe that long-term modeling is important, i.e., reducing the temporal support in ST-E2E leads to sub-optimal performance. With a 32GB GPU memory limit, ST-E2E with a maximum temporal support of 12 seconds achieves 4.1\% lower average mAP than \ModelName~with 32 second temporal support. We also point out that the ST-E2E variant that spans $32$ seconds using sparsely sampled frames (i.e., $8$ vs. $15$ fps) also produces $2.9\%$ worse performance than \ModelName. We observe similar trends for the models trained under the 12GB GPU memory constraint.
Additionally, our results indicate that the the end-to-end training of a short-term feature extractor is also important as LT-Frozen baseline achieves 6.5\% lower accuracy than \ModelName. We observe this trend in both 12GB and 32GB GPU memory settings.

Overall, we can conclude that \ModelName~achieves the best accuracy-memory trade-offs under both 12GB and 32GB GPU memory constraints. Specifically, \ModelName~outperforms the LT-Frozen and ST-E2E baselines by a large margin especially with tighter GPU memory constraints (i.e. 12GB).

\subsection{Comparison to the State-of-the-Art}

Next, we compare \ModelName~to the state-of-the-art methods as shown in Tab.~\ref{tab:compare_sota}. The upper part of Tab.~\ref{tab:compare_sota} includes methods that operate on pre-extracted action recognition features. The middle section of Tab.~\ref{tab:compare_sota} includes recently proposed end-to-end trainable methods, AFSD and DaoTAD, that operate on small spatial video resolutions (i.e., $96\times 96$ and $112\times112$ respectively) to fit into GPU memory. Lastly, in the bottom part of the table, we include our \ModelName, which can be trained end-to-end on long $224 \times 224$ videos. 

We experiment with three variants of our method. First, we introduce a variant, named \ModelName-12, which is cheap enough to fit in a 12GB memory GPU, with two backbones I3D and Swin-B for fair comparison with prior methods. Additionally, for the first variant (using I3D backbone), we use the clip sampling rate $r=0.4$ on THUMOS14 and $r=1/3$ on AcitivityNet-1.3, whereas for the latter variant (using Swin-B backbone) the clip sampling rate is set to $r=0.15$ for THUMOS14 and $r=1/8$ for ActivityNet-1.3. Lastly, our best performing variant is \ModelName-32~with Swin-B as its backbone, which uses a clip sampling-rate of $r=0.4$ for THUMOS and $r=0.375$ for ActivityNet. We set these sampling rates so that our model would fit in the available GPU memory.

\begin{table}[t]
\fontsize{8}{10}\selectfont
\setlength{\tabcolsep}{2.3pt}
    \centering
    \captionsetup{font=small}
    \caption{Comparison to the state-of-the-art on THUMOS14 and ActivityNet-v1.3. \textbf{FE} and \textbf{E2E} denote the feature extractor backbone and whether a method is end-to-end trainable respectively. The feature extractor backbones include TS~\cite{simonyan2014two}, I3D~\cite{carreira2017quo}, P3D~\cite{qiu2017learning} and Swin-B~\cite{liu2021video} (denoted as SW). \textbf{Flow} and \textbf{Ext. Cls.} denote whether each method uses optical flow as input and whether an external action recognition classifier is needed respectively. Note that AFSD relies on an external classifier on ActivityNet-1.3.
    }
    \begin{tabular*}{\textwidth}{ccccccccccccccc}
    \toprule
     Method & FE & E2E & Flow  & \multirow{2}{*}{\makecell{Ext. \\ Cls.}} & \multicolumn{8}{c}{mAP} & Mem  \\
      \cmidrule(lr){6-13}
    & & & & & \multicolumn{4}{c}{THUMOS14} & \multicolumn{4}{c}{ActivityNet-1.3} & (GB) \\
      \cmidrule(lr){6-9} \cmidrule(lr){10-13}
    & & & & & 0.3 & 0.5 & 0.7 & Avg.& 0.5 &0.75 & 0.95 & Avg. & \\
      \midrule
     BSN\cite{lin2018bsn} & TS& \xmark & \cmark & \cmark  & 53.5 & 36.9 & 20.0&36.8&46.5&30.0&8.0& 30.0 &-\\
    BMN\cite{lin2019bmn} & TS & \xmark & \cmark  & \cmark  & 56.0&38.8&20.5&38.5&50.1&34.8&8.3&33.9&- \\
    BC-GNN\cite{bai2020boundary} & TS & \xmark & \cmark  & \cmark &57.1&40.4&23.1&40.2&50.6&34.8&\textbf{9.4}&34.3&- \\
    BU-TAL\cite{zhao2020bottom} & I3D & \xmark & \cmark  & \cmark & 53.9&45.4&28.5&43.3&43.5&33.9&9.2&30.1&- \\
    GTAN\cite{long2019gaussian} & P3D & \xmark & \cmark & \cmark & 57.8 & 38.8&-&-&52.6&34.1&8.9&34.3&-\\
    G-TAD\cite{xu2020g} &TS& \xmark & \cmark & \cmark & 54.5&40.2&23.4&39.3&50.4&34.6&9.0&34.1& - \\
    TAL \cite{chao2018rethinking} &I3D& \xmark & \cmark & \xmark &53.2&42.8&20.8&39.8&38.2&18.3&1.3&20.2 & - \\
    RTD-Action~\cite{tan2021relaxed} & TS & \xmark & \cmark & \cmark & 68.3 & 51.9 & 23.7 & 49.0 & 47.2 & 30.7& 8.6 & 30.8 & -\\
    VSGN~\cite{zhao2021video} & TS & \xmark & \cmark & \cmark & 66.7 & 52.4 & 30.4 & 50.2 & 52.4 & 36.0 & 8.4 & 35.1 & -\\
        \midrule
    AFSD\cite{lin2021learning} &I3D& \cmark & \cmark & - & 67.3 &55.5&31.1&52.0 & 52.4&35.3&6.5& 34.4 & 12 \\
    DaoTAD\cite{wang2021rgb} &I3D& \cmark & \xmark & \xmark & 62.8 &53.8&30.1&50.0 & - & - & - & - & 11 \\
    DaoTAD\cite{wang2021rgb} &SW& \cmark & \xmark & \xmark & 72.7 & 59.8 & 33.3 & 56.3 & - & - & - & - & 30 \\
    \cmidrule(lr){1-14}
    \ModelName-12 & I3D &\cmark & \xmark & \xmark & 68.4& 57.6& 30.8& 53.9 & 41.3&27.3&6.3&27.2& 12\\
    \ModelName-12 & SW &\cmark &  \xmark & \xmark & \textbf{76.1}& 63.1 & 34.2& 59.0& 51.4&34.0&7.6&33.7 & 12\\
    \ModelName-32 & SW & \cmark &  \xmark & \xmark & 76.0&\textbf{63.2}&\textbf{34.5}&\textbf{59.2} & \textbf{54.1}& \textbf{36.2}&7.9&\textbf{35.6}& 29 \\
    \bottomrule
    \end{tabular*}
    \label{tab:compare_sota}
\end{table}

\textbf{THUMOS14.} The results in Tab.~\ref{tab:compare_sota} (the left part of the table), indicate several interesting trends. First, we notice that despite using a small spatial resolution, the end-to-end trainable methods such as AFSD and DaoTAD, outperform methods that operate on pre-extracted action recognition features by a large margin. Second, our results indicate that the memory-constrained \ModelName-12 with an I3D backbone outperforms a strong AFSD baseline by a substantial margin according to all evaluation metrics. Moreover, when increasing the GPU memory constraints, \ModelName-12 achieves 7.2\% higher accuracy on average than AFSD. We note that the GPU consumption for \ModelName~is 29GB, which is still within the capacity of the mainstream Tesla V100 GPUs. We also point out that even when using the same amount of GPU memory as prior methods, our method still largely outperforms previous SOTAs, i.e., \ModelName-12 with I3D backbone and VSwin-B backbone outperforms AFSD by 1.9\% and 7.0\%.

\textbf{ActivityNet.} First, we point out that all the previous methods achieve strong TAL results on ActivityNet while relying on an external action recognition classifier~\cite{xiong2016cuhk}, which ensembles the predictions from ResNet-200 and Inception-V2 models operating on RGB and optical flow inputs. Instead, we simplify this pipeline by predicting action boundaries and categories using a single model. Not only is our proposed framework simpler and more efficient, but it also outperforms all previous approaches by 1.2\% using RGB inputs alone. One interesting observation is that \ModelName~with I3D backbone is 6.5\% lower than \ModelName~with a Swin-B backbone. Our analysis of this result reveals that \ModelName-12 variant with an I3D backbone achieves a low video-level action recognition accuracy (78.2\%) while the accuracy of \ModelName-12 with Swin-B backbone is 90.1\%. We also note that the accuracy of an external action recognition classifier \cite{xiong2016cuhk} used by AFSD is 88.9\%. This empirical finding also explains why all previous methods require an external action recognition classifier.

\textbf{HACS.}
We train \ModelName~with Swin-B as backbone using the same network structure and hyperparameters as in AcitivityNet-1.3. We report these results in Tab.~\ref{tab:hacs}. These results suggest that similar to our previously considered datasets, \ModelName~also achieves state-of-the-art results on the HACS-Segment dataset. Specifically, it outperforms the GTAD~\cite{xu2020g} and BMN~\cite{lin2019bmn,qing2021temporal} baselines by $9.1\%$ and $0.7\%$ average mAP respectively without an external classifier.

\begin{table}[tbp]
\setlength{\tabcolsep}{6pt}
	\centering
	\caption{Our results (in mAP) on the HACS-Segment dataset.}
	\label{tab:hacs}

	\begin{tabular}{c c c c c}
		\toprule
		Backbone & \multicolumn{4}{c}{mAP(\%)} \\
		\cmidrule(lr){2-5}
	& 0.5 & 0.75 & 0.95 & Avg.\\
		\midrule
		GTAD~\cite{xu2020g}&-&-&-&27.5\\
		BMN~\cite{lin2019bmn,qing2021temporal}&52.5&\textbf{36.4}&10.4&35.8\\
		\ModelName&\textbf{55.0}&36.1&\textbf{11.8}&\textbf{36.5}\\
		\bottomrule
	\end{tabular}
\end{table}

\textbf{Inference Speed Discussion.} Compared with previous methods, we use a larger backbone and higher spatial resolution. On the other hand, our proposed framework is much simpler than the frameworks of many prior TAL methods. 
In particular, we use a single model with only RGB frames as input while most previous methods adopt a two-stream approach that requires an external action classifier. This is costly, because (i) optical flow extraction is slow and because (ii) training and inference of an external action classifier is also time-consuming.

Due to the complexity of the existing systems, and the lack of publicly available implementations, it is difficult to quantitatively measure the inference speed of many prior methods. However, we note that in general, \ModelName~provides a much simpler, elegant and more efficient framework to the TAL problem.
For example, consider performing inference on AcitivyNet-1.3 using a RTX A6000 GPU. The overall inference speed of a recent state-of-the-art AFSD~\cite{lin2021learning} is 11.74s/video, which includes (i) optical flow extraction, (ii) processing two modalities and (iii) performing video-level action classification using~\cite{xiong2016cuhk}. On the other hand, \ModelName~only costs 1.58s/video while outperforming AFSD by 1.2\% mAP.

\begin{table}[tbp]
\small
 \captionsetup{font=small}
 \caption{Ablation studies on THUMOS14: \textbf{(a)} \ModelName~works well even for a small sampling rate $r$;
    \textbf{(b)} The temporal consistency module leads to $1.5\%$ boost in the average mAP; 
    \textbf{(c)} \ModelName~performs better with longer temporal support. For \textbf{(a)} the backbone is Swin-B with spatial resolution $224\times224$. For \textbf{(b)} and \textbf{(c)}, the backbone is Swin-T with spatial resolution $112\times 112$.
} 
    \centering
    \begin{subtable}[t]{.33\linewidth}
        \label{tbl:abl_a}
        \centering
        \captionsetup{width=.75\textwidth,font=small}
        \caption{Analysis of the clip sampling rate $r$.}
        \begin{tabular}{ccc}
        \toprule
        $r$ & mAP(\%) & Mem \\
        \midrule
        0.15 & 59.0 & 12 \\
        0.3 & 59.4 &  22 \\
        0.4 & 59.2 & 29 \\
        0.6 & \textbf{60.0} & 45 \\
        1.0 & - & OOM \\
        \bottomrule
        \end{tabular}
    \end{subtable}
    \begin{subtable}[t]{.31\linewidth}
        \label{tbl:abl_b}
        \centering
        \captionsetup{width=.75\textwidth,font=small}
        \caption{Importance of Temporal Consistency Module (TCM).}
        \begin{tabular}{cc}
        \toprule
        TCM & mAP(\%) \\
        \midrule
        \xmark & 51.1\\
        \cmark & \textbf{52.6} \\
        \bottomrule
        \end{tabular}
    \end{subtable}
    \begin{subtable}[t]{.31\linewidth}
        \label{tbl:abl_c}
        \centering
        \captionsetup{width=.75\textwidth,font=small}
        \caption{Temporal Support (TS) analysis.}
        \begin{tabular}{cc}
		\toprule
		TS (sec)&mAP(\%)\\
		\midrule
        8 & 41.5 \\
        16 & 49.7 \\
        24 & 51.5 \\
        32 & 52.8 \\
        40 & \textbf{53.3} \\
		\bottomrule
        \end{tabular}
    \end{subtable}
    \label{tab:ablation_study}
\end{table}

\subsection{Ablation Study}

Lastly, we study various design choices of our \ModelName~model. Specifically, we investigate (i) TAL performance as a function of our clip sampling rate $r$, (ii) the importance of the temporal consistency module (TCM) and (iii) TAL performance as a function of temporal support. We present these results below.

\textbf{Accuracy vs. Clip Sampling Rate.} During training, our model samples a fraction of $r$ total short-term clips from an untrimmed video input. We study the performance as a function of clip sampling rate $r$. From Tab.~\ref{tab:feature_extraction}(a), we can observe that i) GPU memory usage is proportional to the sampling rate $r$; ii) standard end-to-end training ($r=1$) causes out-of-memory (OOM) error; iii) \ModelName~performs quite well even with a very small sampling rate $0.15$, i.e., the TAL accuracy in mAP drops by only 1.0\% while reducing the GPU memory usage to only $12$ GB (compared to $>45$GB using $r=1.0$)

\textbf{Importance of Temporal Consistency Module (TCM).} As shown in Tab.~\ref{tab:ablation_study}(b), our proposed TCM increases the average mAP by 1.5\%, which indicates its importance to our overall framework. More conceptually, these results suggest that encouraging long-range interactions between the memory features, and the online-processed features can alleviate the feature distribution inconsistency issue, which is also suggested by the visualized features before and after TCM in Appendix.~\ref{a_tcm}.

\textbf{Analysis of Temporal Support.} We evaluate \ModelName~when using different temporal support (measured in seconds). Based on the results in Tab~\ref{tab:ablation_study}(c), we observe that longer temporal supports leads to consistently higher mAP.


%% file: conclusions.tex
\section{Discussion}
We present \ModelName, a long-memory Transformer for temporal action localization. Our method is simple, flexible, and it can be efficiently trained on long high-resolution videos for TAL. Furthermore, we demonstrate that \ModelName~significantly outperforms previous TAL approaches on the THUMOS14 and ActivityNet-1.3 benchmarks. 


Some readers might wonder whether optimizing the GPU memory usage for long-video processing is a valuable contribution since modern GPUs can accommodate larger and larger GPU memory requirements. Furthermore, there exist many prior memory saving techniques such as Gradient Checkpoint~\cite{chen2016training} and Mixed Precision~\cite{micikevicius2017mixed}.
Despite the advances in GPU hardware, and new developments in  memory saving techniques, we believe that \ModelName~is still a valuable contribution to the research community. With the new developments in GPU hardware, the demands for higher resolution video analysis and larger models also grow. Thus, such demands pose new GPU memory-related challenges, especially for long-term video understanding tasks such as temporal action localization. We also note that \ModelName~can be easily combined with the existing memory-saving techniques, which we demonstrated in our experiments. Our future work involves extending our framework to various multimodal settings that involve processing both visual inputs and language.

%% file: appendix.tex
\section{Implementation Details}

Here, we provide more details related to the (i) long memory module and (ii) temporal boundary localization module of our \ModelName~model.

\subsection{Long Memory Module}
The implementation details of the proposed Long Memory Module (LMM) and Temporal Consistency Module (TCM) are as shown in Alg.~\ref{alg:lmm}. The inputs are first participate into $N_c$ clips. Among these clips, we sample $N_s$ clips to be processed by the Short-term Transformer Encoder and the remaining $N_c -N_s$ clips by LMM. The clip features extracted by the encoder is also used to update the LMM. All the clips features are fed to the TCM to generate more consistent features. The output features of TCM are the input to the temporal boundary localization module.

\begin{algorithm}[t]
\caption{Pseudocode of short-term feature extraction and feature sampling from long-term memory. }
\label{alg:lmm}
\definecolor{codeblue}{rgb}{0.25,0.5,0.5}
\lstset{
  backgroundcolor=\color{white},
  basicstyle=\ttfamily\footnotesize,
  columns=fullflexible,
  breaklines=true,
  captionpos=b,
  commentstyle=\fontsize{8pt}{8pt}\color{codeblue},
  keywordstyle=\fontsize{8pt}{8pt},
}
\begin{lstlisting}[language=python]
    # encoder: short-term Transformer encoder.
    # clips: video clips (N_c x L_c x H x W x 3).
    # long_memory: the pre-extracted features for this video (N_c x L_f x C_f).
    # r: sampling rate (float).

    # sample clips processed by encoder
    sampled_idx = uniform_sample(N_c, r)
    remaining_idx = [idx for idx in range(N_c) if idx not in sampled_idx]
    sampled_clips = clips[sampled_idx]

    # Short-term Transformer Encoder
    sampled_features = encoder.forward(sampled_clips) # shape: [N_s, L_f, C_f]

    # Long-term Memory Module
    mem_features = long_memory[remaining_idx]. # shape: [N_c-N_s, L_f, C_f]
    long_memory[sampled_idx] = sampled_features.detach()

    # Temporal Consistent Module
    ## gather features
    features = zeros(N_c,*sampled_features.shape[1:])
    features[sampled_idx] = sampled_features
    features[remaining_idx] = mem_features
    features = features.reshape(N_c*L_f, C_f) #shape: [N_c*L_f, C_f]
    ## refine features
    for i in range(L):
        features = TransformerLayer(features) #shape: [N_c*L_f, C_f]
\end{lstlisting}
\end{algorithm}

\subsection{Temporal Boundary-Localization Module}
\label{a_tblm}
Given the refined features $f_r \in \mathbb{R}^{C_c \times L}$, the Temporal Boundary-Localization Module (TBLM) aims to produce the action boundaries and categories for each action instance. We use different TBLMs for THUMOS14~\cite{idrees2017thumos} and ActivityNet~\cite{caba2015activitynet}.

\begin{figure}[tb]
	\centering
	\includegraphics[width=0.98\textwidth]{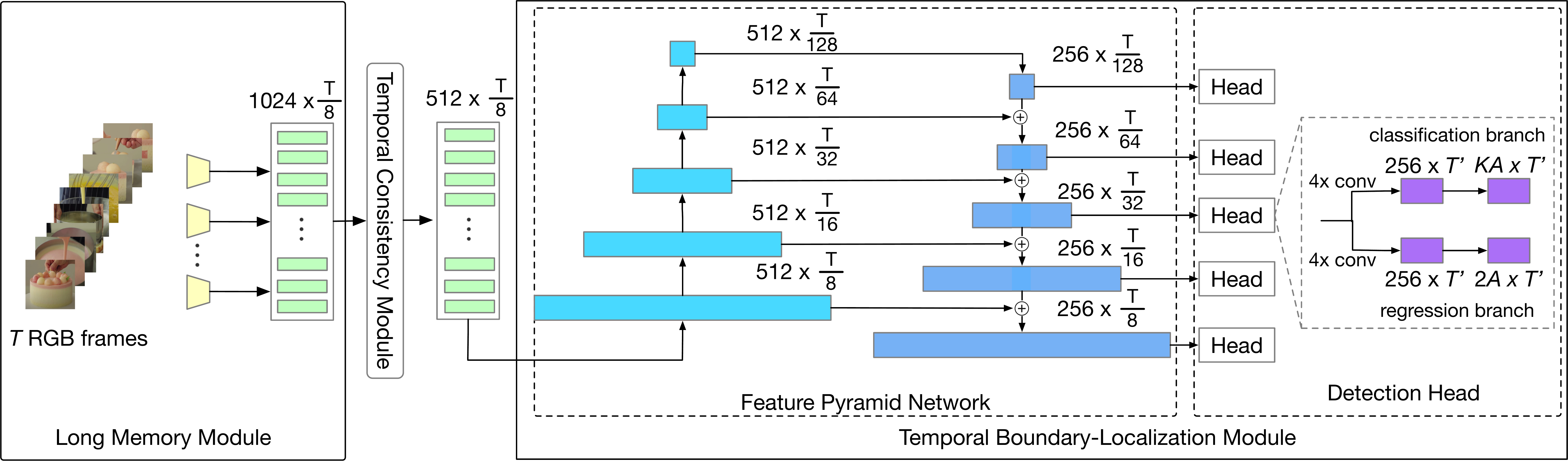}
	\caption{Network structure for THUMOS14.}
	\label{fig:structure-thumos}
\end{figure}

\textbf{THUMOS14.} The detailed architecture is shown in Fig.~\ref{fig:structure-thumos}. The TBLM is composed of a Feature-Pyramid Network (FPN) and a Detection Head. The Detection Head is taken from DaoTAD~\cite{wang2021rgb}. In the FPN, the features are downsampled (bottom to up) using 1D kernel-3, stride-2 convolutions and are upsampled (up to bottom) by linear interpolation along the temporal dimension. We use Focal loss \cite{lin2017focal} for the sigmoid-activated classification branch and DIoU loss \cite{zheng2020distance} for the regression branch. The weights are 1 for both losses. We refer readers to \cite{wang2021rgb} for more details.

\begin{figure}[tb]
	\centering
	\includegraphics[width=0.98\textwidth]{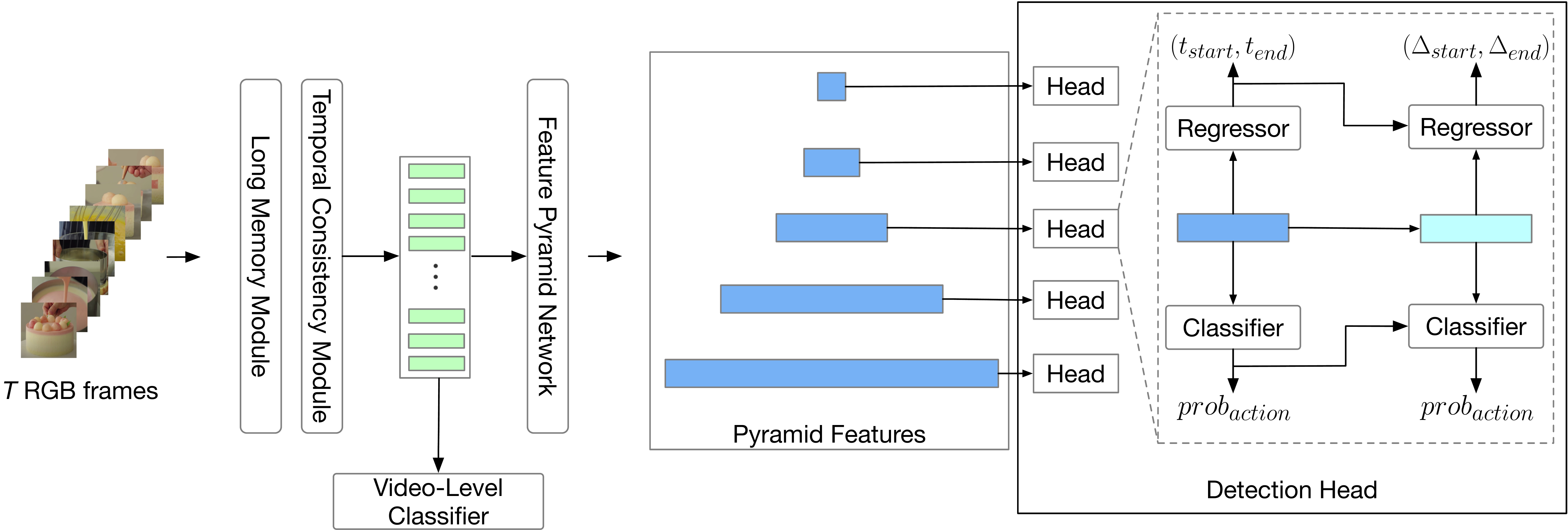}
	\caption{Network structure for ActivityNet-1.3. The same Long Memory Module, Temporal Consistency Module and Feature Pyramid Network are used as in THUMOS14.}
	\label{fig:structure-anet}
\end{figure}

\textbf{ActivityNet-1.3.} The detailed architecture is shown in Fig.~\ref{fig:structure-anet}. We use the same Long Memory Module, Temporal Consistency Module, Feature Pyramid Network as in THUMOS14. We adopt the Detection Head design from AFSD \cite{lin2021learning}.
Additionally, after the Temporal Consistency Module, we also add a video-level classifier composed of a global average pooling layer, dropout layer with drop-rate 0.5 and a linear layer with a dimensionality equal to the number of action classes.
AFSD Detection Head is a two-stage detector. First, it uses a Basic Prediction Module to predict the coarse action boundaries and action-agnostic classes (background or not).
Then a Saliency-based Refinement Module is used to refine the predicted boundaries and action-agnostic classes. Finally, we assign each predicted action proposal with the action category predicted by the video-level classifier.
We use Cross-entropy loss for video-level classifier, Focal loss \cite{lin2017focal} for classification branches in the detection head, tIoU loss \cite{lin2021learning} for the regression of Basic Prediction Module and L1 loss for the boundary refinement in the Saliency-based Prediction Module.
The weights are 1 for all the losses.
We refer the readers to \cite{lin2021learning} for more specific details related to the Detection Head. 

\section{Additional Results}

\subsection{Importance of Temporal Consistency Module}
\label{a_tcm}
\begin{figure}[tb]
	\centering
	\includegraphics[width=0.98\textwidth]{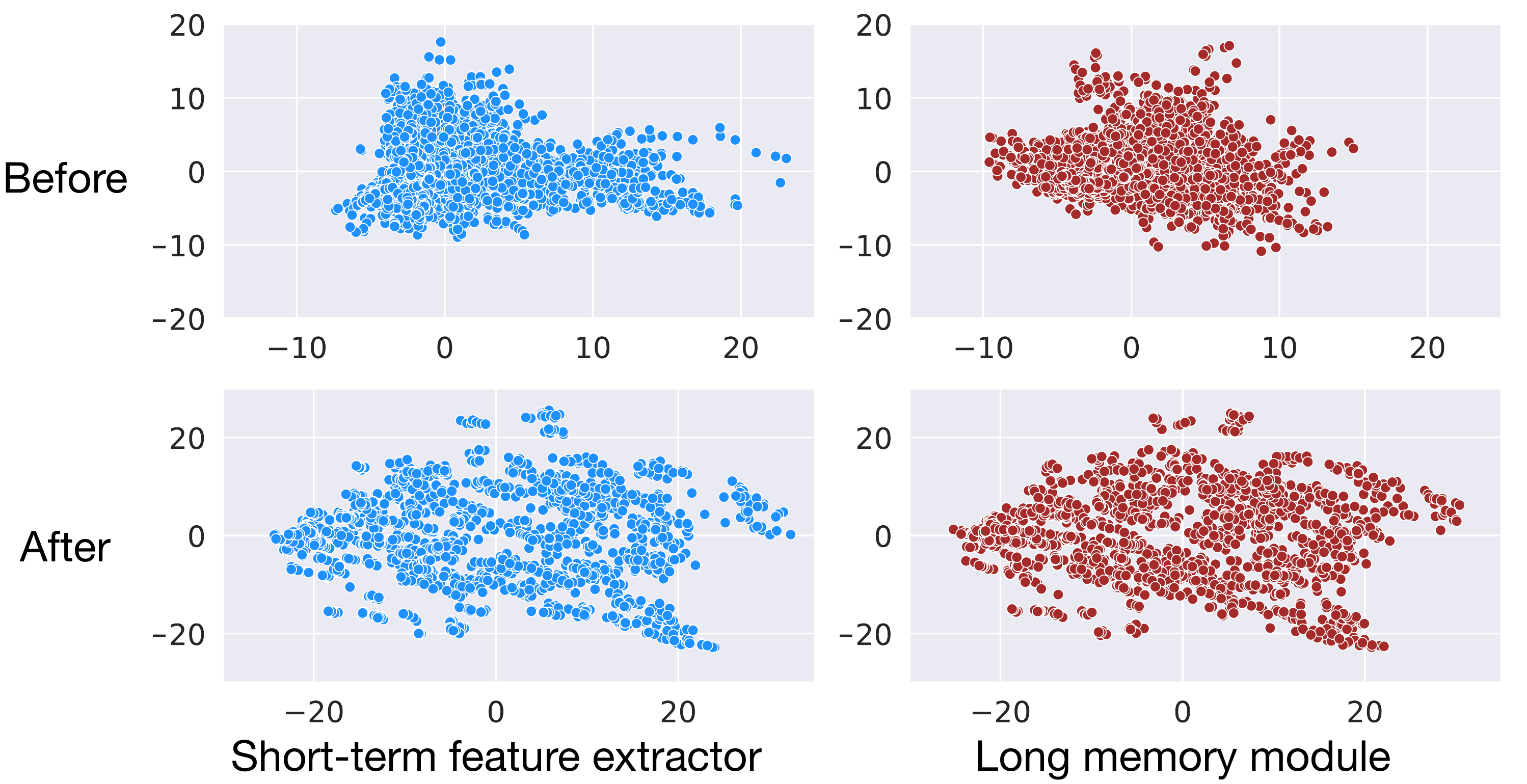}
	\caption{Network structure for THUMOS14.}
	\label{fig:tcm}
\end{figure}

In addition to the quantitative results in the main paper, we visualize the features before and after Temporal Consistency Module (TCM) as in Fig. we extracted four sets of features: features from short-term feature extractor (1) before, and (2) after the TCM, and features from the Long Memory Module (3) before, and (4) after the TCM. We then applied PCA and plotted the first two principal components as shown Fig.~\ref{fig:tcm}. We observe that the features from the short-term feature extractor and long-term memory are more similar after the TCM than they were before the TCM. This suggests that TCM effectively reduces the inconsistency between features from the short-term feature extractor and long memory module.

\begin{table}[tbhp]
\setlength{\tabcolsep}{6pt}
	\centering
	\caption{Ablating different short-term transformer encoders within our \ModelName~framework on THUMOS14~\cite{idrees2017thumos}.}
	\label{tab:encoder_ablation}

	\begin{tabular}{ccccccc}
		\toprule
		Transformer Encoder &\multicolumn{6}{c}{mAP(\%)}\\
	    \cmidrule(lr){2-7}
		&0.3&0.4&0.5&0.6&0.7&Avg.\\
		\midrule
		Swin-T~\cite{liu2021video}&72.7 & 69.0 & 60.8 & 48.3 & 34.3 & 57.0 \\
        Swin-S~\cite{liu2021video}& 74.9 & 70.3 & 62.1 & 48.9 & 34.3 & 58.1 \\
        Swin-B~\cite{liu2021video}&76.0 & 71.5 & 63.2 & 50.9 & 34.5 & 59.2 \\
		\bottomrule
	\end{tabular}
	\vspace{-0.5cm}
\end{table}

\subsection{Ablating Different Short-term Transformer Encoders}

The flexibility of our \ModelName~model allows us to use any short-term transformer encoder as our clip-level backbone. To demonstrate \ModelName's generalization with different backbones, we experiment with different variations of Swin Transformers~\cite{liu2021video}, i.e. Swin-tiny, Swin-small and Swin-base. As shown in Tab.~\ref{tab:encoder_ablation}, \ModelName~achieves pretty high average mAPs on all the backbones.

\subsection{Ablating Temporal Support}
Due to long actions (e.g., $30$ seconds in length), our model needs to span long temporal extent. Thus, here, we evaluate \ModelName when using different temporal support (measured in seconds). Based on the results in Tab~\ref{tab:temporal-support}, we observe that longer temporal supports leads to consistently higher average mAP. 

\begin{table}[tbhp!]
\setlength{\tabcolsep}{6pt}
	\centering
    \caption{Temporal Support (TS) ablation on THUMOS14~\cite{idrees2017thumos}. The model is \ModelName~\cite{wang2021rgb} with Swin-T as backbone and spatial resolution $112\times 112$. We observe that longer temporal supports leads to higher average mAP.}
	\label{tab:temporal-support}

	\begin{tabular}{ccccccc}
		\toprule
		TS (sec)&\multicolumn{6}{c}{mAP(\%)}\\
		\cmidrule(lr){2-7}
		&0.3&0.4&0.5&0.6&0.7&Avg.\\
		\midrule
        8 & 59.8 & 54.1 & 45.4 & 31.3 & 17.0 & 41.5 \\
        16 & 65.0 & 60.3 & 52.5 & 42.6 & 28.3 & 49.7 \\
        24 & 65.7 & 62.0 & 53.8 & 45.0 & 31.0 & 51.5 \\
        32 & 66.9 & 63.2 & \textbf{56.7} & \textbf{46.0} & 31.1 & 52.8 \\
        40 & \textbf{68.0} & \textbf{63.7} & 56.2 & \textbf{46.0} & \textbf{32.6} & \textbf{53.3} \\
		\bottomrule
	\end{tabular}
\end{table}

%% file: main.bbl
\begin{thebibliography}{10}
\providecommand{\url}[1]{\texttt{#1}}
\providecommand{\urlprefix}{URL }
\providecommand{\doi}[1]{https://doi.org/#1}

\bibitem{bagchi2021hear}
Bagchi, A., Mahmood, J., Fernandes, D., Sarvadevabhatla, R.K.: Hear me out:
  Fusional approaches for audio augmented temporal action localization. arXiv
  preprint arXiv:2106.14118  (2021)

\bibitem{bai2020boundary}
Bai, Y., Wang, Y., Tong, Y., Yang, Y., Liu, Q., Liu, J.: Boundary content graph
  neural network for temporal action proposal generation. In: European
  Conference on Computer Vision. pp. 121--137. Springer (2020)

\bibitem{bertasius2021space}
Bertasius, G., Wang, H., Torresani, L.: Is space-time attention all you need
  for video understanding. arXiv preprint arXiv:2102.05095  \textbf{2}(3), ~4
  (2021)

\bibitem{caba2015activitynet}
Caba~Heilbron, F., Escorcia, V., Ghanem, B., Carlos~Niebles, J.: Activitynet: A
  large-scale video benchmark for human activity understanding. In: Proceedings
  of the ieee conference on computer vision and pattern recognition. pp.
  961--970 (2015)

\bibitem{carreira2017quo}
Carreira, J., Zisserman, A.: Quo vadis, action recognition? a new model and the
  kinetics dataset. In: proceedings of the IEEE Conference on Computer Vision
  and Pattern Recognition. pp. 6299--6308 (2017)

\bibitem{chao2018rethinking}
Chao, Y.W., Vijayanarasimhan, S., Seybold, B., Ross, D.A., Deng, J.,
  Sukthankar, R.: Rethinking the faster r-cnn architecture for temporal action
  localization. In: Proceedings of the IEEE Conference on Computer Vision and
  Pattern Recognition. pp. 1130--1139 (2018)

\bibitem{chen2016training}
Chen, T., Xu, B., Zhang, C., Guestrin, C.: Training deep nets with sublinear
  memory cost. arXiv:1604.06174  (2016)

\bibitem{cheng2022stochastic}
Cheng, F., Xu, M., Xiong, Y., Chen, H., Li, X., Li, W., Xia, W.: Stochastic
  backpropagation: A memory efficient strategy for training video models. In:
  Proceedings of the IEEE/CVF Conference on Computer Vision and Pattern
  Recognition. pp. 8301--8310 (2022)

\bibitem{choromanski2020rethinking}
Choromanski, K., Likhosherstov, V., Dohan, D., Song, X., Gane, A., Sarlos, T.,
  Hawkins, P., Davis, J., Mohiuddin, A., Kaiser, L., et~al.: Rethinking
  attention with performers. arXiv preprint arXiv:2009.14794  (2020)

\bibitem{fan2021multiscale}
Fan, H., Xiong, B., Mangalam, K., Li, Y., Yan, Z., Malik, J., Feichtenhofer,
  C.: Multiscale vision transformers. In: Proceedings of the IEEE/CVF
  International Conference on Computer Vision. pp. 6824--6835 (2021)

\bibitem{feichtenhofer2020x3d}
Feichtenhofer, C.: X3d: Expanding architectures for efficient video
  recognition. In: Proceedings of the IEEE/CVF Conference on Computer Vision
  and Pattern Recognition. pp. 203--213 (2020)

\bibitem{feichtenhofer2019slowfast}
Feichtenhofer, C., Fan, H., Malik, J., He, K.: Slowfast networks for video
  recognition. In: Proceedings of the IEEE/CVF international conference on
  computer vision. pp. 6202--6211 (2019)

\bibitem{gao2020accurate}
Gao, J., Shi, Z., Wang, G., Li, J., Yuan, Y., Ge, S., Zhou, X.: Accurate
  temporal action proposal generation with relation-aware pyramid network. In:
  Proceedings of the AAAI Conference on Artificial Intelligence. vol.~34, pp.
  10810--10817 (2020)

\bibitem{gao2018ctap}
Gao, J., Chen, K., Nevatia, R.: Ctap: Complementary temporal action proposal
  generation. In: Proceedings of the European conference on computer vision
  (ECCV). pp. 68--83 (2018)

\bibitem{goyal2017something}
Goyal, R., Ebrahimi~Kahou, S., Michalski, V., Materzynska, J., Westphal, S.,
  Kim, H., Haenel, V., Fruend, I., Yianilos, P., Mueller-Freitag, M., et~al.:
  The" something something" video database for learning and evaluating visual
  common sense. In: Proceedings of the IEEE international conference on
  computer vision. pp. 5842--5850 (2017)

\bibitem{he2020momentum}
He, K., Fan, H., Wu, Y., Xie, S., Girshick, R.: Momentum contrast for
  unsupervised visual representation learning. In: Proceedings of the IEEE/CVF
  conference on computer vision and pattern recognition. pp. 9729--9738 (2020)

\bibitem{he2016deep}
He, K., Zhang, X., Ren, S., Sun, J.: Deep residual learning for image
  recognition. In: Proceedings of the IEEE conference on computer vision and
  pattern recognition. pp. 770--778 (2016)

\bibitem{hendrycks2016gaussian}
Hendrycks, D., Gimpel, K.: Gaussian error linear units (gelus). arXiv preprint
  arXiv:1606.08415  (2016)

\bibitem{huang2016deep}
Huang, G., Sun, Y., Liu, Z., Sedra, D., Weinberger, K.Q.: Deep networks with
  stochastic depth. In: European conference on computer vision. pp. 646--661.
  Springer (2016)

\bibitem{idrees2017thumos}
Idrees, H., Zamir, A.R., Jiang, Y.G., Gorban, A., Laptev, I., Sukthankar, R.,
  Shah, M.: The thumos challenge on action recognition for videos “in the
  wild”. Computer Vision and Image Understanding  \textbf{155},  1--23 (2017)

\bibitem{jiang2019stm}
Jiang, B., Wang, M., Gan, W., Wu, W., Yan, J.: Stm: Spatiotemporal and motion
  encoding for action recognition. In: Proceedings of the IEEE/CVF
  International Conference on Computer Vision. pp. 2000--2009 (2019)

\bibitem{kwon2020motionsqueeze}
Kwon, H., Kim, M., Kwak, S., Cho, M.: Motionsqueeze: Neural motion feature
  learning for video understanding. In: European Conference on Computer Vision.
  pp. 345--362. Springer (2020)

\bibitem{lin2021learning}
Lin, C., Xu, C., Luo, D., Wang, Y., Tai, Y., Wang, C., Li, J., Huang, F., Fu,
  Y.: Learning salient boundary feature for anchor-free temporal action
  localization. In: Proceedings of the IEEE/CVF Conference on Computer Vision
  and Pattern Recognition. pp. 3320--3329 (2021)

\bibitem{lin2019tsm}
Lin, J., Gan, C., Han, S.: Tsm: Temporal shift module for efficient video
  understanding. In: Proceedings of the IEEE/CVF International Conference on
  Computer Vision. pp. 7083--7093 (2019)

\bibitem{lin2019bmn}
Lin, T., Liu, X., Li, X., Ding, E., Wen, S.: Bmn: Boundary-matching network for
  temporal action proposal generation. In: Proceedings of the IEEE/CVF
  International Conference on Computer Vision. pp. 3889--3898 (2019)

\bibitem{lin2017single}
Lin, T., Zhao, X., Shou, Z.: Single shot temporal action detection. In:
  Proceedings of the 25th ACM international conference on Multimedia. pp.
  988--996 (2017)

\bibitem{lin2018bsn}
Lin, T., Zhao, X., Su, H., Wang, C., Yang, M.: Bsn: Boundary sensitive network
  for temporal action proposal generation. In: Proceedings of the European
  conference on computer vision (ECCV ). pp. 3--19 (2018)

\bibitem{lin2017focal}
Lin, T.Y., Goyal, P., Girshick, R., He, K., Doll{\'a}r, P.: Focal loss for
  dense object detection. In: Proceedings of the IEEE international conference
  on computer vision. pp. 2980--2988 (2017)

\bibitem{liu2020progressive}
Liu, Q., Wang, Z.: Progressive boundary refinement network for temporal action
  detection. In: Proceedings of the AAAI Conference on Artificial Intelligence.
  vol.~34, pp. 11612--11619 (2020)

\bibitem{liu2019multi}
Liu, Y., Ma, L., Zhang, Y., Liu, W., Chang, S.F.: Multi-granularity generator
  for temporal action proposal. In: Proceedings of the IEEE/CVF Conference on
  Computer Vision and Pattern Recognition. pp. 3604--3613 (2019)

\bibitem{liu2021swin}
Liu, Z., Lin, Y., Cao, Y., Hu, H., Wei, Y., Zhang, Z., Lin, S., Guo, B.: Swin
  transformer: Hierarchical vision transformer using shifted windows. In:
  Proceedings of the IEEE/CVF International Conference on Computer Vision. pp.
  10012--10022 (2021)

\bibitem{liu2021video}
Liu, Z., Ning, J., Cao, Y., Wei, Y., Zhang, Z., Lin, S., Hu, H.: Video swin
  transformer. arXiv preprint arXiv:2106.13230  (2021)

\bibitem{long2019gaussian}
Long, F., Yao, T., Qiu, Z., Tian, X., Luo, J., Mei, T.: Gaussian temporal
  awareness networks for action localization. In: Proceedings of the IEEE/CVF
  Conference on Computer Vision and Pattern Recognition. pp. 344--353 (2019)

\bibitem{micikevicius2017mixed}
Micikevicius, P., Narang, S., Alben, J., Diamos, G., Elsen, E., Garcia, D.,
  Ginsburg, B., Houston, M., Kuchaiev, O., Venkatesh, G., et~al.: Mixed
  precision training. arXiv preprint arXiv:1710.03740  (2017)

\bibitem{patrick2021keeping}
Patrick, M., Campbell, D., Asano, Y., Misra, I., Metze, F., Feichtenhofer, C.,
  Vedaldi, A., Henriques, J.F.: Keeping your eye on the ball: Trajectory
  attention in video transformers. Advances in Neural Information Processing
  Systems  \textbf{34} (2021)

\bibitem{qing2021temporal}
Qing, Z., Su, H., Gan, W., Wang, D., Wu, W., Wang, X., Qiao, Y., Yan, J., Gao,
  C., Sang, N.: Temporal context aggregation network for temporal action
  proposal refinement. In: Proceedings of the IEEE/CVF Conference on Computer
  Vision and Pattern Recognition. pp. 485--494 (2021)

\bibitem{qiu2017learning}
Qiu, Z., Yao, T., Mei, T.: Learning spatio-temporal representation with
  pseudo-3d residual networks. In: proceedings of the IEEE International
  Conference on Computer Vision. pp. 5533--5541 (2017)

\bibitem{shou2017cdc}
Shou, Z., Chan, J., Zareian, A., Miyazawa, K., Chang, S.F.: Cdc:
  Convolutional-de-convolutional networks for precise temporal action
  localization in untrimmed videos. In: Proceedings of the IEEE conference on
  computer vision and pattern recognition. pp. 5734--5743 (2017)

\bibitem{simonyan2014two}
Simonyan, K., Zisserman, A.: Two-stream convolutional networks for action
  recognition in videos. Advances in neural information processing systems
  \textbf{27} (2014)

\bibitem{su2020bsn++}
Su, H., Gan, W., Wu, W., Qiao, Y., Yan, J.: Bsn++: Complementary boundary
  regressor with scale-balanced relation modeling for temporal action proposal
  generation. arXiv preprint arXiv:2009.07641  (2020)

\bibitem{tan2021relaxed}
Tan, J., Tang, J., Wang, L., Wu, G.: Relaxed transformer decoders for direct
  action proposal generation. In: Proceedings of the IEEE/CVF International
  Conference on Computer Vision. pp. 13526--13535 (2021)

\bibitem{tran2015learning}
Tran, D., Bourdev, L., Fergus, R., Torresani, L., Paluri, M.: Learning
  spatiotemporal features with 3d convolutional networks. In: Proceedings of
  the IEEE international conference on computer vision. pp. 4489--4497 (2015)

\bibitem{tran2018closer}
Tran, D., Wang, H., Torresani, L., Ray, J., LeCun, Y., Paluri, M.: A closer
  look at spatiotemporal convolutions for action recognition. In: Proceedings
  of the IEEE conference on Computer Vision and Pattern Recognition. pp.
  6450--6459 (2018)

\bibitem{vaswani2017attention}
Vaswani, A., Shazeer, N., Parmar, N., Uszkoreit, J., Jones, L., Gomez, A.N.,
  Kaiser, {\L}., Polosukhin, I.: Attention is all you need. Advances in neural
  information processing systems  \textbf{30} (2017)

\bibitem{wang2021rgb}
Wang, C., Cai, H., Zou, Y., Xiong, Y.: Rgb stream is enough for temporal action
  detection. arXiv preprint arXiv:2107.04362  (2021)

\bibitem{wang2018appearance}
Wang, L., Li, W., Li, W., Van~Gool, L.: Appearance-and-relation networks for
  video classification. In: Proceedings of the IEEE conference on computer
  vision and pattern recognition. pp. 1430--1439 (2018)

\bibitem{wang2016temporal}
Wang, L., Xiong, Y., Wang, Z., Qiao, Y., Lin, D., Tang, X., Gool, L.V.:
  Temporal segment networks: Towards good practices for deep action
  recognition. In: European conference on computer vision. pp. 20--36. Springer
  (2016)

\bibitem{wang2018temporal}
Wang, L., Xiong, Y., Wang, Z., Qiao, Y., Lin, D., Tang, X., Van~Gool, L.:
  Temporal segment networks for action recognition in videos. IEEE transactions
  on pattern analysis and machine intelligence  \textbf{41}(11),  2740--2755
  (2018)

\bibitem{wang2020linformer}
Wang, S., Li, B.Z., Khabsa, M., Fang, H., Ma, H.: Linformer: Self-attention
  with linear complexity. arXiv preprint arXiv:2006.04768  (2020)

\bibitem{wang2020multi}
Wang, X., Gao, C., Zhang, S., Sang, N.: Multi-level temporal pyramid network
  for action detection. In: Chinese Conference on Pattern Recognition and
  Computer Vision (PRCV). pp. 41--54. Springer (2020)

\bibitem{wu2019long}
Wu, C.Y., Feichtenhofer, C., Fan, H., He, K., Krahenbuhl, P., Girshick, R.:
  Long-term feature banks for detailed video understanding. In: Proceedings of
  the IEEE/CVF Conference on Computer Vision and Pattern Recognition. pp.
  284--293 (2019)

\bibitem{xiong2016cuhk}
Xiong, Y., Wang, L., Wang, Z., Zhang, B., Song, H., Li, W., Lin, D., Qiao, Y.,
  Van~Gool, L., Tang, X.: Cuhk \& ethz \& siat submission to activitynet
  challenge 2016. arXiv preprint arXiv:1608.00797  (2016)

\bibitem{xu2020g}
Xu, M., Zhao, C., Rojas, D.S., Thabet, A., Ghanem, B.: G-tad: Sub-graph
  localization for temporal action detection. In: Proceedings of the IEEE/CVF
  Conference on Computer Vision and Pattern Recognition. pp. 10156--10165
  (2020)

\bibitem{xu2021long}
Xu, M., Xiong, Y., Chen, H., Li, X., Xia, W., Tu, Z., Soatto, S.: Long
  short-term transformer for online action detection. Advances in Neural
  Information Processing Systems  \textbf{34},  1086--1099 (2021)

\bibitem{you2022megan}
You, C., Han, L., Feng, A., Zhao, R., Tang, H., Fan, W.: Megan: Memory enhanced
  graph attention network for space-time video super-resolution. In:
  Proceedings of the IEEE/CVF Winter Conference on Applications of Computer
  Vision. pp. 1401--1411 (2022)

\bibitem{you2022class}
You, C., Zhao, R., Liu, F., Chinchali, S., Topcu, U., Staib, L., Duncan, J.S.:
  Class-aware generative adversarial transformers for medical image
  segmentation. arXiv preprint arXiv:2201.10737  (2022)

\bibitem{zeng2019graph}
Zeng, R., Huang, W., Tan, M., Rong, Y., Zhao, P., Huang, J., Gan, C.: Graph
  convolutional networks for temporal action localization. In: Proceedings of
  the IEEE/CVF International Conference on Computer Vision. pp. 7094--7103
  (2019)

\bibitem{zhang2021temporal}
Zhang, C., Gupta, A., Zisserman, A.: Temporal query networks for fine-grained
  video understanding. In: Proceedings of the IEEE/CVF Conference on Computer
  Vision and Pattern Recognition. pp. 4486--4496 (2021)

\bibitem{zhang2018s3d}
Zhang, D., Dai, X., Wang, X., Wang, Y.F.: S3d: single shot multi-span detector
  via fully 3d convolutional networks. arXiv preprint arXiv:1807.08069  (2018)

\bibitem{zhao2021video}
Zhao, C., Thabet, A.K., Ghanem, B.: Video self-stitching graph network for
  temporal action localization. In: Proceedings of the IEEE/CVF International
  Conference on Computer Vision. pp. 13658--13667 (2021)

\bibitem{zhao2019hacs}
Zhao, H., Torralba, A., Torresani, L., Yan, Z.: Hacs: Human action clips and
  segments dataset for recognition and temporal localization. In: Proceedings
  of the IEEE/CVF International Conference on Computer Vision. pp. 8668--8678
  (2019)

\bibitem{zhao2020bottom}
Zhao, P., Xie, L., Ju, C., Zhang, Y., Wang, Y., Tian, Q.: Bottom-up temporal
  action localization with mutual regularization. In: European Conference on
  Computer Vision. pp. 539--555. Springer (2020)

\bibitem{zhao2017temporal}
Zhao, Y., Xiong, Y., Wang, L., Wu, Z., Tang, X., Lin, D.: Temporal action
  detection with structured segment networks. In: Proceedings of the IEEE
  International Conference on Computer Vision. pp. 2914--2923 (2017)

\end{thebibliography}
